\documentclass[10pt, conference, compsocconf]{IEEEtran}
\usepackage{blindtext, url}
\usepackage[english]{babel}
\usepackage{graphicx}
\usepackage{float}
\usepackage[font=small]{caption}

\ifCLASSINFOpdf
\else
\fi
\begin{document}

%
\title{Autonomous Fault Detection in Self-Healing Systems \\using Restricted Boltzmann Machines}



%

\author{\IEEEauthorblockN{Chris Schneider}
\IEEEauthorblockA{School of Computer Science\\
University of St Andrews\\
Scotland, UK\\
chris.schneider@st-andrews.ac.uk}
\and
\IEEEauthorblockN{Adam Barker}
\IEEEauthorblockA{School of Computer Science\\
University of St Andrews\\
Scotland, UK\\
adam.barker@st-andrews.ac.uk}
\and
\IEEEauthorblockN{Simon Dobson}
\IEEEauthorblockA{School of Computer Science\\
University of St Andrews\\
Scotland, UK\\
simon.dobson@st-andrews.ac.uk}
}


\maketitle

\begin{abstract}
Autonomously detecting and recovering from
faults is one approach for reducing  the operational complexity
and  costs associated with managing computing environments.
We present a novel methodology for autonomously generating
investigation leads that help identify systems faults, and extends our previous work in this area by leveraging Restricted Boltzmann Machines (RBMs) and contrastive divergence learning to analyse changes in historical feature data. This allows us   to heuristically identify the root cause of a fault, and demonstrate an improvement to the state of the art by showing feature data can be predicted heuristically beyond a single instance to include entire sequences of information.

\end{abstract}


%
\IEEEpeerreviewmaketitle

\section{Introduction}
 The operational costs of large-scale computing environments are continuing to increase.
        In order to address this problem, self-managing systems are being developed that reduce the supervisory needs of computing environments.
        Self-healing systems are one such  example, and operate by autonomously detecting then recovering from faults.
        Although there have been numerous advances in both of these aspects, most self-healing systems continue to require periodic human oversight \cite{schneider2013asurveyofselfhealingsystemsframeworks,dobson2010fulfillingthevisionofautonomiccomputing,psaier2010asurveyonselfhealingsystemsapproachesandsystems,mccann2004evaluationIssuesinAutonomicComputing}.
        This constraint poses challenges for the continued reduction of costs, and restricts self-healing recovery strategies to reactive approaches \cite{dean2012ublunsupervisedbehaviorlearningforpredictingperformanceanomaliesinvirtualizedcloudsystems}.
The importance of reducing human oversight in managing computing environments is multi-faceted.
        Although numerous direct benefits exist--such as the reduction staff involvement and their associated operating costs--further achievements can also be realised.
        Notably, self-healing systems have properties that are showing inherent benefits to change control schemas, and preserving baseline configurations \cite{miorandi2010embryonicmodelsforselfhealingdistributedservices}.
 
The lack of change control or a baseline configuration can both introduce faults and present problems in determining their respective sources.
        Additionally, self-healing systems methodologies are also showing the capability to both detect and resolve faults without human supervision \cite{dean2012ublunsupervisedbehaviorlearningforpredictingperformanceanomaliesinvirtualizedcloudsystems,ramirez2011platoageneticalgorithmapproachtoruntimereconfigurationsinautonomiccomputingsystems,shehory2007aselfhealingapproachtodesigninganddeployingcomplexdistributedandconcurrentsoftwaresystems}. 
        This is important when considering operational constraints-- such as costs and time requirements associated with training technical members of staff--to achieve these same results.
        If a system can find an appropriate recovery solution without the need for a subject matter expert, the associated costs can be immediately recovered.
        
However, achieving these goals is non-trivial and has posed notable challenges in both Machine Learning, and Artificial/Computational Intelligence.
        There is no assurance, for example, that self-healing systems leveraging evolutionary or search-space algorithms will find an appropriate solution for a given fault, or that any solution found will be optimal.
        Furthermore, computational costs of approaches that leverage these methodologies are typically higher than others, and inherently carry a certain amount of risk of failing to identify or resolve faults.
        Anecdotal evidence suggests that in professional computing environments the failure to recognise or mitigate a fault is never an acceptable state.
        It is clear, however, that such circumstances do happen under human supervision, and they may be inevitable. 
        The fact remains that moving to a software based approach poses challenges and questions regarding accountability--currently associated with human administrators--and liability.
        Both of these topics are outside of the scope of this paper, but the preference in supervised management approaches lends evidence to the desirability of these criteria \cite{cardellini2011mosesaframeworkforqosdrivenruntimeadaptationofserviceorientedsystems,li2013aselfhealingframeworkforqosawarewebservicecompositionviacasebasedreasoning,menasce2011sassyaframeworkforselfarchitectingserviceorientedsystems,rilling2006vignetowardsaselfhealinggridoperatingsystem,schuler2004scalablepeertopeerprocessmanagementtheosirisapproach,stojnic2012osirissrasafetyringforselfhealingdistributedcompositeservice}.
The question remains: 
        How can we further the autonomous behaviours of self-healing systems whilst reducing the operating costs of large-scale computing environments?
        
        Previous research has shown that it is possible to synthesise new, valid systems configurations \cite{ramirez2011platoageneticalgorithmapproachtoruntimereconfigurationsinautonomiccomputingsystems}, and determine common relationships between features \cite{zheng20123dimensionalrootcausediagnosisviacoanalysis,garvin2013failureavoidanceinconfigurablesystemsthroughfeaturelocality}.
        This has helped to reactively build recovery solutions in an unsupervised fashion and predict the validity of specific systems' configurations, respectively.
        The ability to autonomously identify anomalies has also been demonstrated by using a special type of unsupervised artificial neural network (ANN) \cite{dean2012ublunsupervisedbehaviorlearningforpredictingperformanceanomaliesinvirtualizedcloudsystems} called a  self-organising map  \cite{kohonen1990theselforganisingmap}, and in our previous work using Hidden Markov Models (HMMs) \cite{schneider2014autonomousfaultdetectioninselfhealingsystemscomparinghiddenmarkovmodelsandartificialneuralnetworks}. 
        These approaches emphasise predictive behaviours by leveraging historical configuration data collected from a local system. 
However, at present there are no  performance evaluations of self-healing systems utilising these methodologies. In order to understand how effective these approaches are they must be compared. 
                
        In this paper we extend our previous approach for autonomously evaluating the source of a fault within a system
by using Restricted Boltzmann Machines (RBMs) to predict the
state of a feature, show how that prediction can be used to identify the source of a fault via a comparison between the expected and actual result, and illustrate how a series of features can be synthesised given an input vector. Using this approach allows the application rather than an administrator to find the specific cause of an anomaly.
A comparison is provided  with the previous results using ether HMMs or ANNs, 
and the respective advantages and performance metrics are discussed.

Lastly, we conclude with groundwork for potentially discovering new avenues to identify faults via more robust analysis in feature locality.
It is our intent to  continue to develop this research further and to eventually demonstrate potential reductions in the cost of operating large-scale IT environments through automation. 

The rest of this paper is organised as follows:
        Section 2 contains details of the approach.
        Sections 3 and 4 describe the implementation, and key components of the methodology, respectively. Section 5 presents some early experimental results whilst  Section 6 concludes with some directions for future
exploration.

\section{Approach}
Using RBMs it is possible to identify the source of faults within a system without human intervention.
        RBMs use a learning algorithm to evaluate and predict changes in feature behaviour by utilising historical performance and configuration data periodically gathered from the system.
        This data is then autonomously classified through the use of fitness tests as either valid or invalid.
        Results from these tests determine the overall state of the system, and subsequently categorise the data
collected in an identical fashion.
 This information is then used to provide direction to the RBM.
        
        If the system passes all of its fitness tests, the associated configuration is assumed to be valid.
        This data is then converted into vectors based on state changes and then saved to disk for potential analysis. 
        As the system passes its fitness tests, more information is added to the existing saved data sets, until a collection is achieved of a pre-determined value. 
        However, as systems behaviour can and is expected to change over time, previously learned information is gradually expired.
        This allows for elasticity in predictions by limiting the information learned to a recent time-series.
           
        If the system does not pass all of its fitness tests, the associated configuration is assumed to be invalid.
        Once an invalid state has been determined an evaluation is done for each feature's behaviours based on the previously information. Rather than using a greedy evaluation--as in previous instantiations--the RBM uses a lazy evaluation strategy on features that are determined to have changed from the last known good configuration and the faulty configuration data.
    
Features that are determined to have behaved in an unexpected manner are added to a list of potential faults, along with a confidence value.
        The confidence value is determined by how unlikely the behaviour is to have occurred according to the RBM.
        Using the confidence value, the list of potential faults is then sorted in descending order.
        This provides both a measure of effectiveness of the application for determining the root cause of the fault, and an ability to prioritise subsequent self-healing strategies.

\section{Implementation}

In order to achieve the aforementioned approach, this experiment leverages C\# and the Windows Management Instrumentation (WMI) framework for data collection.
        A small application periodically interfaces with the WMI service based on a polling interval.
        The polling interval determines two properties: How frequently the WMI framework is to be queried, and how much elasticity to account for in behavioural pattern analysis.
        Although both values are fully adjustable, for the purposes of this experiment the polling interval is set at 60 seconds, and the total size of the dataset collection is limited to 30 samples.
        Each dataset is referenced within a list, and contains a collection of tables that individually correspond to a WMI class.
        As the WMI framework is queried, these tables are populated, associated with their respective dataset, and then categorised.
        Lastly, the information to be gathered is determined at run-time via a dictionary that stores a unique identifier value and the names of the WMI classes to be queried.

        The categorisation of dataset information is accomplished via fitness tests that validate the responsibilities of the virtual machine.
        In this case the virtual machine's primary purpose is to act as a web-server for both internal and external clients.
        Rather than using unit tests to verify a series of specific properties, fitness tests emphasise the validation of high-level processes and functions.
This allows the application rather than an administrator to find the specific cause the anomaly.
Furthermore, the use of fitness tests in this experiment accomplishes three goals:
        1.) It emulates more closely the use of policies than unit tests--a goal for self-managing systems described by prior research \cite{kephart2011autonomiccomputingthefirstdecade,kephart2003thevisionofautonomiccomputing,kephart2004anartificialintelligenceperspectiveonautonomiccomputingpolicies},
         2.) It roughly mirrors standard practice in existing computing environments where operational readiness testing or service-level agreements are required, and
        3.) It establishes the groundwork for feeding in the results of this experiment with planned future research. 
        
        As previously stated, once a dataset is categorised as either valid or invalid the application will either save the collected information to disk for layer analysis, or it will look for

\begin{figure*}[t]
\centering
\includegraphics[scale=0.45]{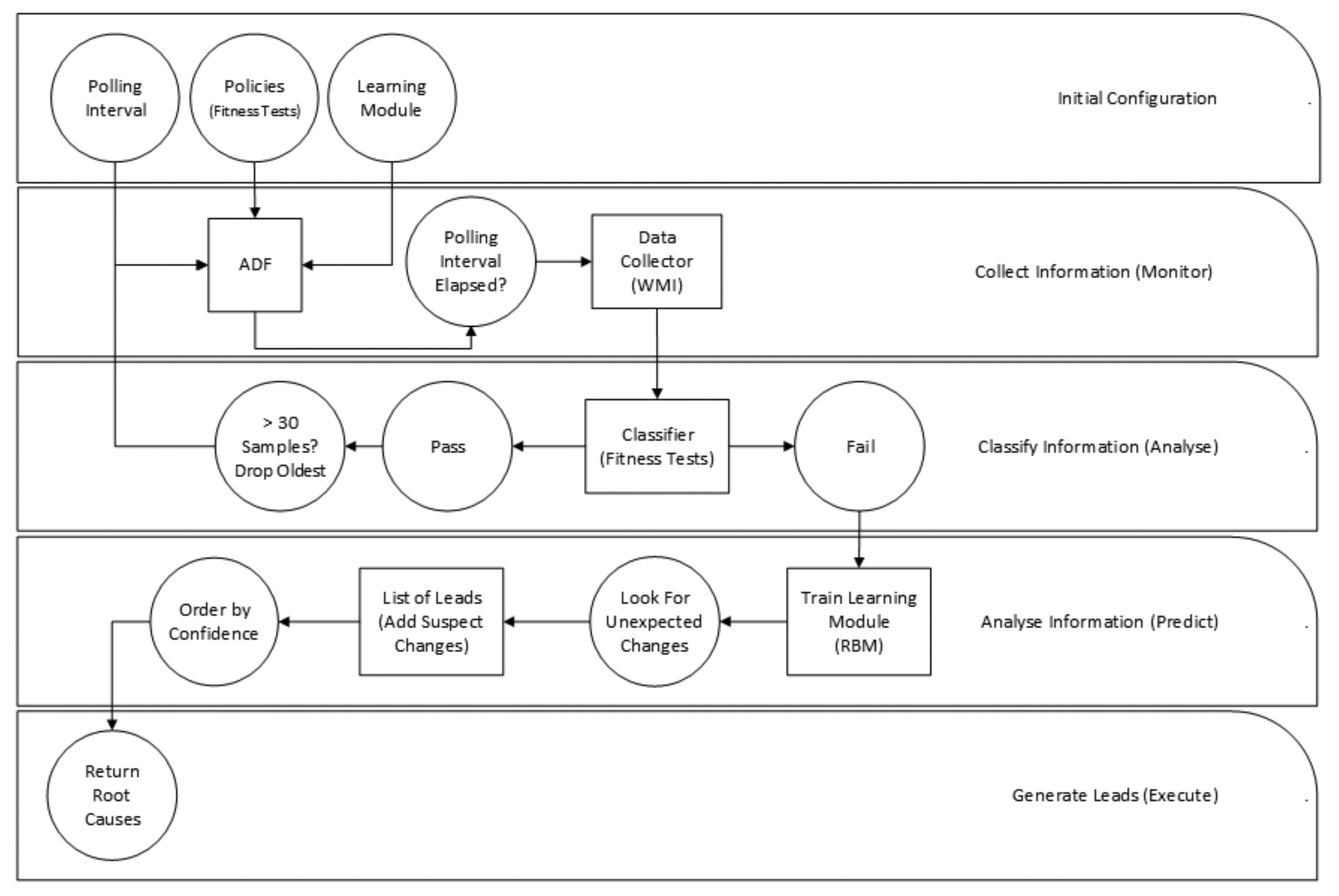}
\caption{Anomaly Detection Framework Logic \& Architecture  Diagram}
\label{anomalydetectionframeworklogicumldiagram}
\end{figure*}

anomalies, respectively.
        The dataset is determined to be valid if it passes all of its fitness tests.
        If this occurs,  each property within the collection of datasets is evaluated against itself.
        The hardest part of this procedure is uniquely identifying the objects that have been queried.
        
        WMI does not provide a unique identifier for the values it produces, so an intersection is used to identify like-objects based on the lowest expected rate of change for a specific value within a given WMI class.
        This value, identified by column, is the primary reason for aforementioned WMI class dictionary's existence.
        After verifying that the application has no more than the maximum number of datasets, any changes--including removed or newly discovered properties--are catalogued and a vector is produced that contains change information.
        It is this vector that is used to autonomously train the anomaly detection framework (ADFs) in this experiment.
        
The ADF in this experiment leverages contrastive divergent learning \cite{carreiraperpinan2002oncontrastivedivergencelearning}.
        This algorithm was chosen due to its suitability with RBMs  object structure, and its ability to both categorise information and synthesise a series of inputs based on an output value. The ADF was  implemented  via the AForge.NET \cite{kirillov2013aforge}  and Accord.NET Frameworks \cite{souza2013accordnet}, including the learning algorithm which is responsible for processing observed feature behaviours into probabilities, and the RBM object code. This code is used in conjunction with the ADF's classification methods for the datasets collected via WMI, and metrics gathering algorithms.  

If the dataset is determined to be invalid, the feature's behaviours are analysed by the ADFs for unexpected changes. Any feature that does not match the ADF's predicted values is added to a list of potential faults, along with a confidence value. As long as the fault source is collected within the WMI data, and the feature behaviours are sufficiently predictable--in this case, any detectable change within 30 samples that is also associated with the fault--the root cause of a fault should be detected by the ADF.

Although other learning algorithms are available, comparing their advantages and disadvantages remain beyond the scope of this experiment. However, this is an  area that hopefully will be explored in the future in a separate publication. Specifically, instead of contrastive divergence learning (CDL), we hope to understand how effective  HMMs utilising the Viterbi algorithm are at generating or predicting sequences of information when compared to CDL \cite{viterbi1967errorboundsforconvulutionalcodesandansymptoticallyoptimumdecodingalgorithm}. The underlying differences between  HMMs and RBMs are not fully explored within this paper as we anticipate readers will be versed in these topics.

\section{Methodology}

This experiment leveraged a virtual machine running Windows 7, Internet Information Services (IIS) 7.5, and one instance of the ADF.
        The virtual machine was cloned from an initial image used in our prior experiments, and consisted of identical base configurations in hardware.
       The hardware itself was unremarkable being a standard image with 1GB of RAM, and a  single disk partition divided into three volumes--one for the OS, the ADF, and the IIS webroot, respectively.
        The software was identical to the original experiment up until the point at which the ADF was allowed to run for a training period of 30 minutes.

        During this time, the fitness tests were evaluated once every 60 seconds.
        If a system passed all of its fitness tests, the ADF would save the configuration it gathered along with an XML schema file to a local data store.
        These files served as a mechanism for loading known good systems configurations quickly and, as a consequence, allowed for more rapid testing.
        Additionally, by approaching the experiment in this fashion we were able to reduce the opportunity for drift in each virtual machines' configuration, and allow for greater reproducibility in the experiment.
Once the machine was trained, it was exposed to either a direct fault injection event  or an adverse configuration change  that was expected to limit the system's ability to either connect to the internet, run IIS-related services, or the ability to access information stored on the the system's disk volumes. 

        The ADF was then responsible for detecting the presence of the fault and generating a potential root cause, as well as reporting on several key attributes including:        
                the total number of true positives,
                        true negatives,
                        false positives,
                        and false negatives, the time taken in ``ElapsedTicks'' from the point in which a fault was detected until the completed generation of the ordered list of potential root causes, and the number of potential root causes (\textit{i.e.} `leads').
        
        True and false positives were determined when a fault was detected and whether or not it was or was not present, respectively.
        Conversely, true and false negatives were determined when a fault was present but not detected. 
        However, due to the nature of false negatives, the number of faults not detected by the application had to be done by hand.
        This was as expected as there was no way, by definition, for the application to detect such a state without external validation.
        It is also the reason that faults in this experiment were injected with the source already being known.
From this information inferential metrics such as precision, time-taken, and leads generated.
        This data was then combined to produce charts showing the performance of the ADFs relative to the same tests.
        
        The type of faults we injected had two variants: Adverse Configuration Changes (ACCs), and Direct Fault Injections (DFIs).
        The former consisted of shutting off services or making changes to the system using normal administrative methods. This included changing disk structures, service states, and other properties that administrators would normally have access to.
        The latter consisted of copying code directly into the address space of another process, which in turn was expected to produce a controlled crash. These faults were introduced in such a way that the fitness tests implemented in the ADF were expected to fail, but it would not be aware of how or why.

        The specific ACCs that were instantiated included: Disabling the network card, disabling the W3SVC service, removing the volume upon which the IIS webroot was contained, removing all free space from any of the three volumes,  disabling network access from one hop above the virtual machine's purview, and  sabotaging the primary DNS resolver entry. The DFIs we instantiated included crashing various services such as: The IIS 7.5 W3SVC service, the Windows IPv4 network stack, and the Windows DNS service.
        Each ACC or DFI was run 6 times using the same ADF which was allowed access to 5, 10, 15, 20, 25, and 30 configuration samples. This allowed us to realise trends within each approach, and to see differences in both output and ADF confidence during each specific test. 

The confidence values for each result were generated using contrastive divergence learning. Once a fault was detected by the ADF, it loaded the sampled data  from disk and instantiated  an individual RBM for any feature that did not have matching historical change data in both the last known good (LKG) and current (\textit{i.e.} faulty) configuration data. The RBM was then trained using the LKG values over 5,000 epochs before attempting to produce two values representing  either an expected or unexpected result against the faulty data series. The highest value was selected as the most likely category of the feature's behaviour, and as a representation of confidence.

The training methodology used in this experiment differs from the previous approach in two key characteristics: It uses lazy evaluation, and a padded series of inputs.
The change from greedy evaluation to lazy was done as a general optimisation; the reasoning for this is explained in further detail in the results section but can be summarised as logical improvements to the framework for skipping features that had identical behavioural data and as an improvement in the total number of computational operations.

In order to train an RBM it is necessary to produce a collection of feature behaviours organised in a series--\textit{i.e.} a matrix of values that
has the same dimensions as the number of samples.
To ensure that this experiment leveraged the same total volume of input data in the previous iteration, and to maintain consistency between the two sets of results, `no data' markers were utilised to complete a series where appropriate. In exchange, the ability to predict a series of values was gained versus only single values in the previous iteration.

\section{Results}

The successful evaluation of this experiment focused on whether or not it was possible to correctly   detect the presence of a fault, and then identify its source using a comparison of  actual and synthesised feature data by leveraging  Restricted Boltzmann Machines. The results from this experiment show that is is possible to meet both of these criteria and this methodology improves upon the previous approach. However, using an RBM  comes with the costs of a longer initial wait time for results, higher variability within those results, and, ideally, larger training sets.

\begin{figure}[t]
\centering
\begin{center}
\includegraphics[width=.5\textwidth]{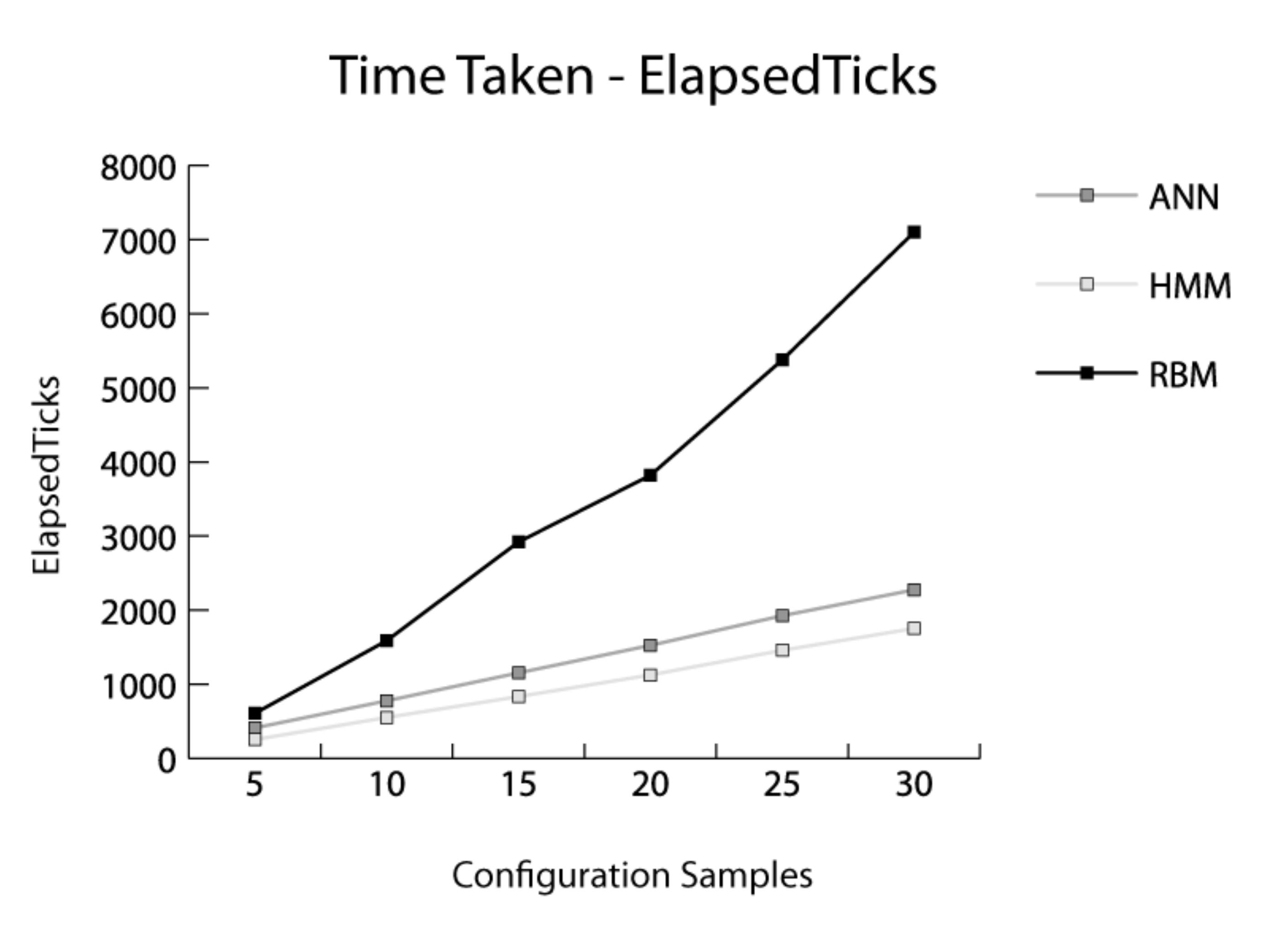}\\
\caption{Time Taken represents the average number of ``ElapsedTicks'' between when a fault was detected and the return of an ordered list of potential root causes based on confidence value.}
\label{rbmtimetaken}
\end{center}
\end{figure}

As expected the RBM required more ElapsedTicks  from the time a fault was detected by the ADF to complete its training and evaluation tasks than the two prior methodologies (Figure \ref{rbmtimetaken}). This was largely due to the aforementioned switch  from a greedy to a lazy evaluation, and the number of epochs used to train  each RBM. In comparison, however, this was an improvement overall in terms of total resources consumed.

In the two previous instances, the  number of total computational cycles used was much higher. Direct observation during runtime showed that both the HMM and ANN ADFs utilised anywhere from 1-3\% of the CPU for approximately 30 seconds for every minute they were active. These cycles were used primarily for WMI data collection and for training the ADFs--both processes exited after 5, and 25 seconds on average, respectively. Since the RBMs were not trained until a fault was explicitly detected,  25 seconds of CPU time  were recovered for each minute the ADF was active.

In instances where a greater number of samples needed to be evaluated, a steep linear increase was observed in the number of ElapsedTicks before the process completed.
This was due to the size of the data collection, which grew faster than  previous approaches.
However, the approach does scale relatively well. Using the modest resources provided to the VM it took about 10 seconds to parse all 30 samples in each iteration--each sample containing up to 30 data points on approximately 6,000 features. Naturally, changing the amount of data collected, or how much total data is stored, will increase the processing time.

Fault position is the primary metric upon which the success of the ADFs were evaluated--it is simply where the correct fault is located within a list of potential root causes. The lower the index value of the correct fault, the better the ADFs overall performance. The ideal ADF returns an index value of 0 every time for the correctly identified fault.
\begin{figure}[t]
\begin{center}
\includegraphics[width=.5\textwidth]{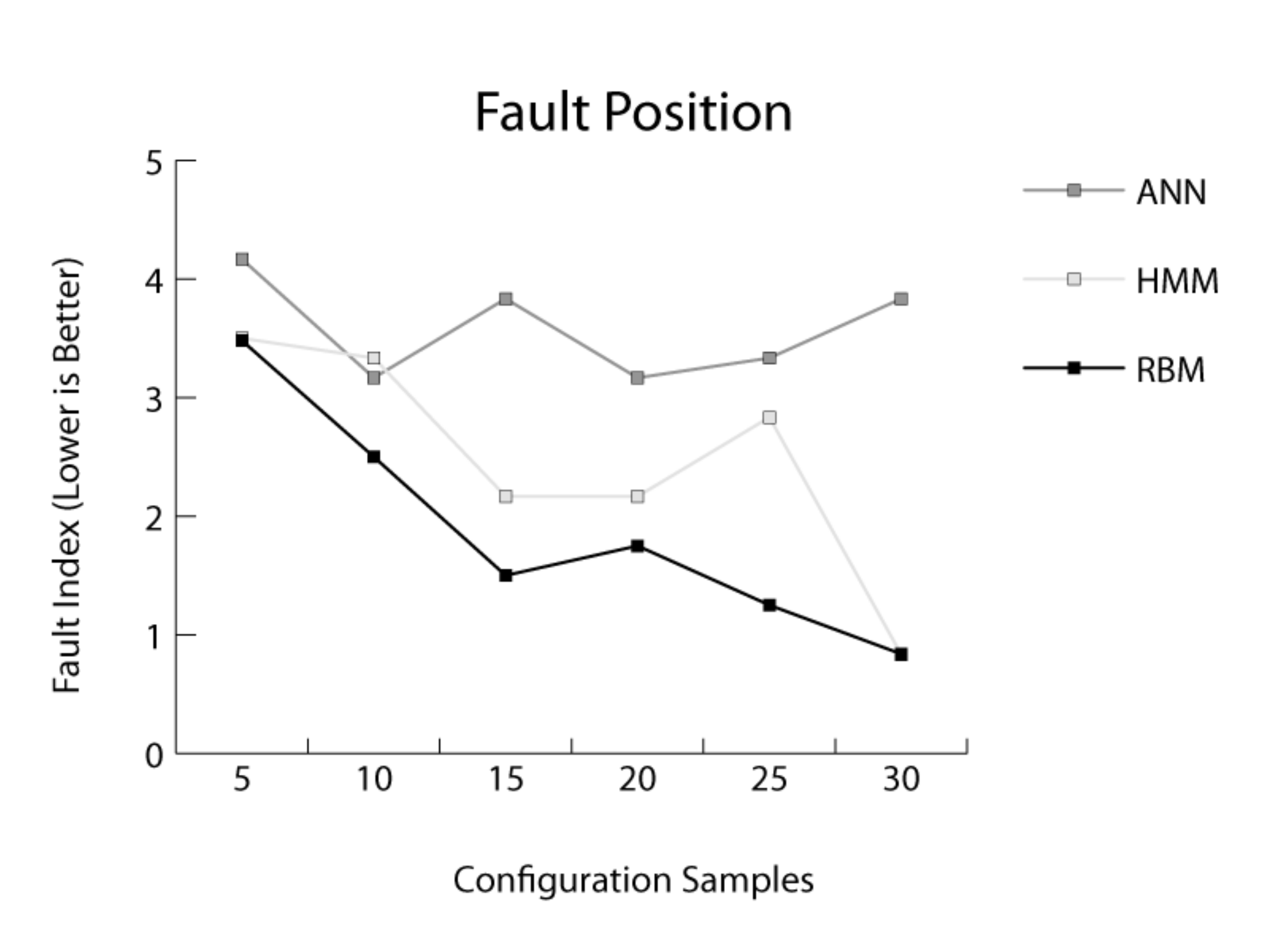}\\
\caption{The average position of the correct fault as returned by the ADF is represented in this graph. Overall the RBM was able to list the correct fault more frequently than the two previous approaches--with an exception at 30 samples.}
\label{rbmfaultposition}
\end{center}
\end{figure}

On average RBM was able to produce   a lower index position for the correct fault than the previous approaches--with the exception of the HMM (Figure \ref{rbmfaultposition}). When using 30 configuration samples the HMM was able to position the correct fault slightly better than the RBM--0.83 versus 0.838, respectively. The gradient of each of these approaches also suggests that the HMM could continue to outpace other strategies.

\begin{figure}[h]
\begin{center}
\includegraphics[width=.5\textwidth]{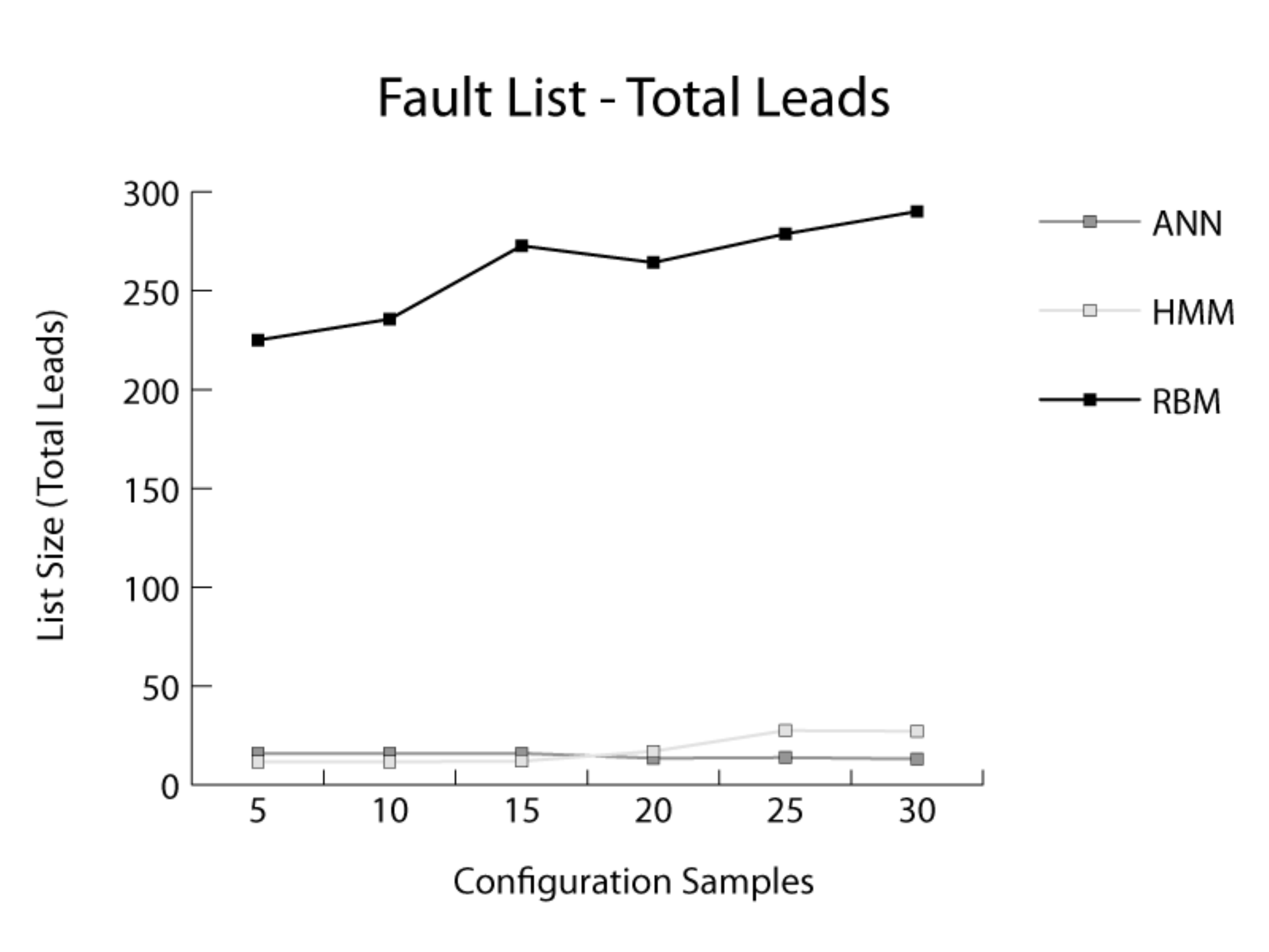}\\
\caption{Each ADF is responsible for generating leads when a fault is detected. This graph represents the average total number of suspect features (\textit{i.e.} `leads') per approach at each sample size.}
\label{rbmleads}
\end{center}
\end{figure}

The number of total leads represents the avenues of exploration that the ADF must account for after each object has been trained, respectively. In the case of the RBM this value was much higher than the two previous approaches. By switching from greedy to lazy evaluation, changes in the system's configuration had to be accounted for all at once when a fault was detected. This meant storing all of the leads over time instead of evaluating them gradually and then accounting for them via the learning algorithm. The result was   growth over time for the total number of features that needed to be evaluated.

\begin{figure}[t]
\begin{center}
\includegraphics[width=.5\textwidth]{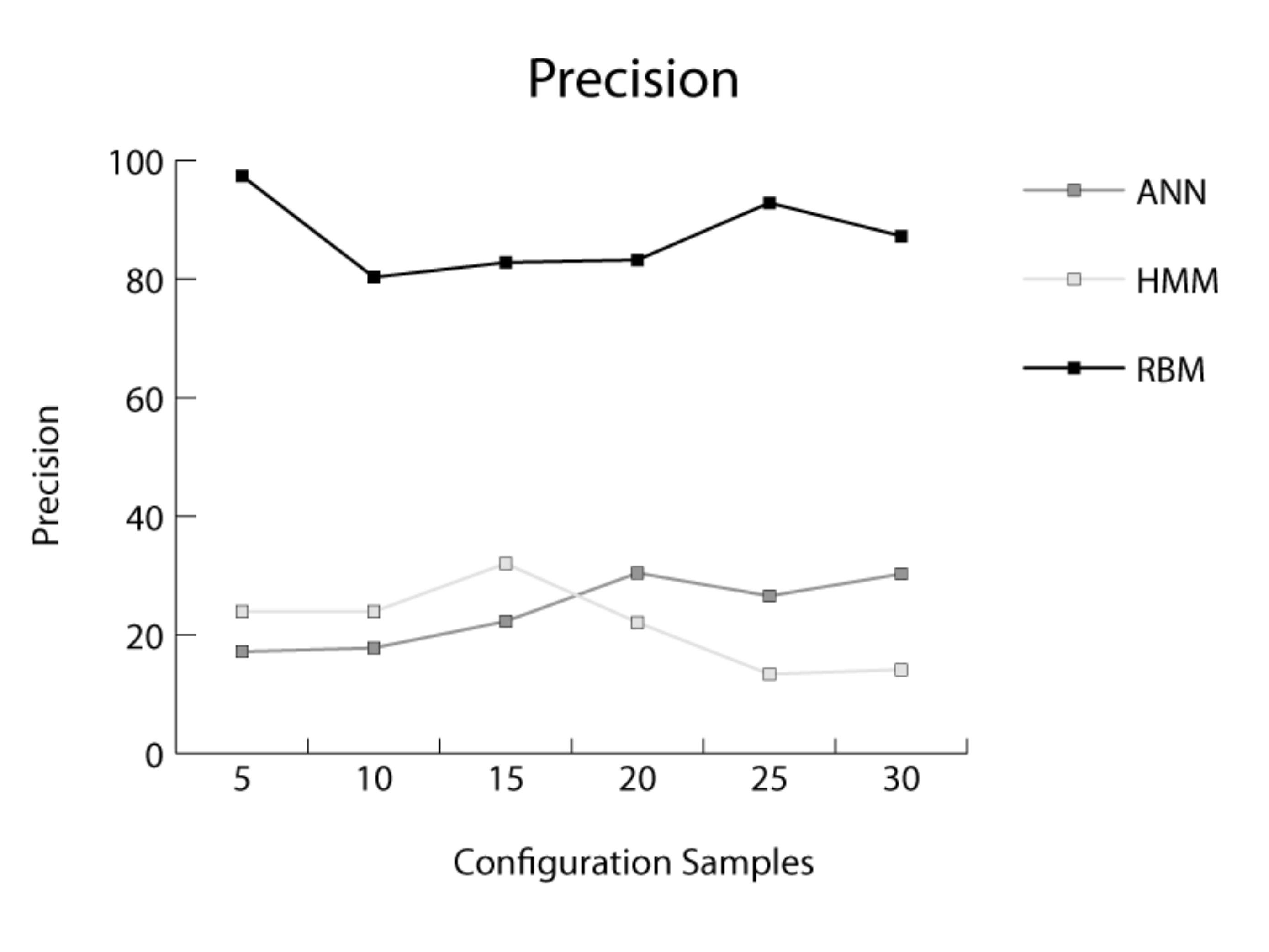} \\
\caption{Precision was measured by taking the total number of correct leads in the list divided by the same value plus the number of leads above these entries.}
\label{rbmprecision}
\end{center}
\end{figure}

\begin{figure}[h]
\centering
\begin{center}
\includegraphics[width=.5\textwidth]{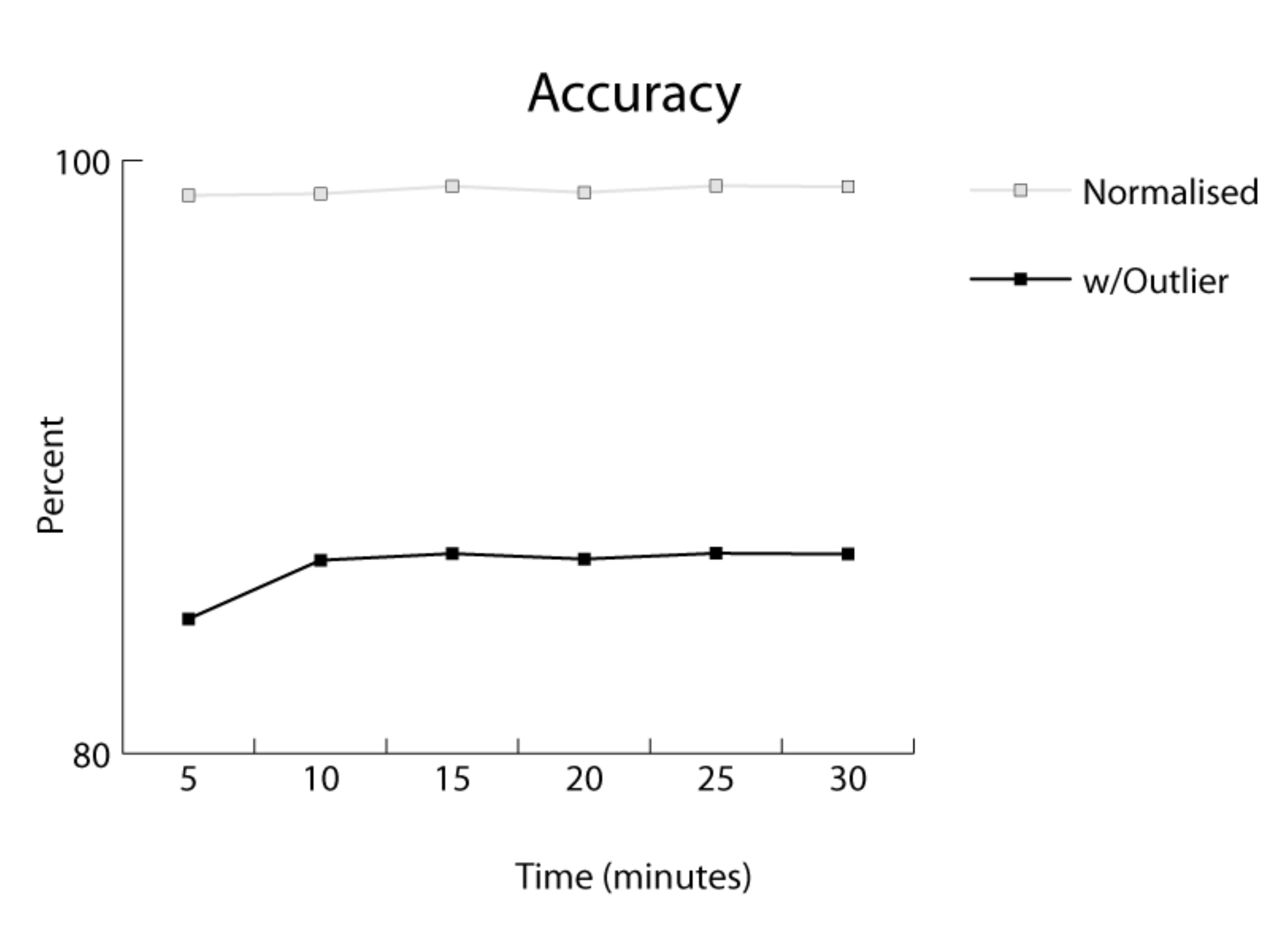}\\
\caption{Accuracy in the RBM ADF showed an improvement over prior approaches. In one instance, however, the correct fault failed to be identified.}
\label{rbmaccuracy}
\end{center}
\end{figure}

Interestingly, the list size did not seem to influence the accuracy of the RBM negatively (Figure \ref{rbmleads}). Based on the fault position, the evaluation of the correct lead was given more precisely and more accurately than previous approaches (Figures ~\ref{rbmprecision} and ~\ref{rbmaccuracy}, respectively). From a  probabilistic perspective, it was expected that there would be a greater amount of variance in which leads were selected.
                 Instead, it was noticed that the results returned by the ADF had a wider range of outputs than previous instances. In the simplest of terms the same inputs would return similar but notably different results.

The variance in the RBM's output seems to be associated with how the RBMs are instantiated. A random seed is used to build each node within the RBM. This value dictates the initial state of the node, and consequently as these values get updated in different order, the paths for each output are assumed to also be somewhat randomised. Since  this information is used to predict entire sequences of data, the chance for a comparison to mismatch seems to manifest at a higher rate than in previous instances. This is both reasonable and expected considering the use of a simulated data set--however, if a full training set were used, we would expect the variance to drop.

In light of this, one possible explanation for the performance improvement is that by keeping all of the potential leads until the end the likelihood of missing the correct feature was lessened. However, this theory has not yet been tested and remains an avenue for future research. By converting both the HMM and ANN ADFs to use lazy evaluation it may be possible to duplicate the results we've seen here with the RBM.

As mentioned in the previous paper, a greater list of leads is not assumed to be better. The ideal ADF will return a list that consists of only the correct avenues for exploration. As such, the approach of the previous ADFs is more likely to meet this criteria than using lazy evaluation in combination with RBMs.

The accuracy of this experiment was  measured by evaluating the number of correct leads in the list that matched the fault source plus the number of leads that didn't match, divided by this same information plus the number of  leads incorrectly categorised above the correct leads, plus the number of false negatives.

Overall the results showed an improvement from previous attempts. In the majority of cases the correct root cause was at  presented within the list of leads--often with a high confidence value. In one instance, however, a test was run to see if the ADF could determine the root cause of a fault that was outside of the local system. This was outside of the scope of the initial experiment, but we were interested in exploring the potential results.

After shutting off a network appliance upstream from the VM,  and using 30 known good configuration samples, the ADF suggested the root cause was a network adapter throughput / speed change. This result was arguably correct--the adapter speed was indeed reduced to zero.
Considering the ADF had no sample data to work with indicating the existing of other potential problems, this result was surprising. With fewer configuration samples the ADF returned an entirely incorrect root cause--the number of total processes running on the system--and thus the accuracy dropped to 14\% for this test.
As such, it has been included here, but with a normalised view that included only the tests we expected to run, and with the outlier which included the test that had a fault outside of the local machine's purview.

Lastly, as in the previous experiment, using fitness functions and a full virtual machine with live input allowed for a direct approach when evaluating the ADF's results. As such, no pre-fabricated model  needed to be provided--the ADFs  built its own expectations of the features' behaviours so long as the fitness tests continued to pass.

\begin{figure}[t]
\centering
\begin{center}
\includegraphics[width=.5\textwidth]{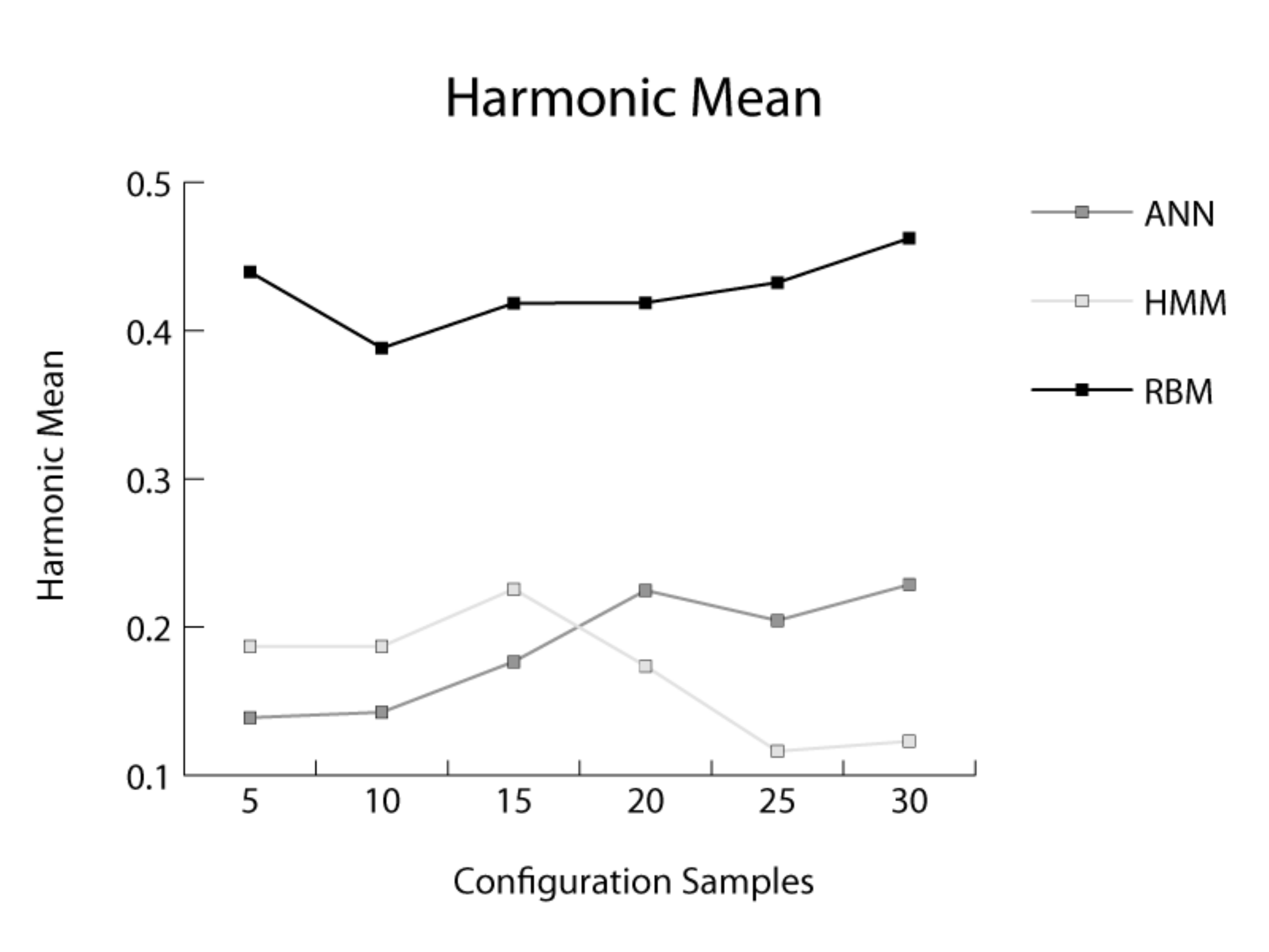}\\
\caption{Harmonic Mean represents the overall performance of the contrastive divergence learning (RBM) algorithm in relation to both Baum-Welch (HMM), and Na\"ive Bayes (ANN).}
\label{rbmharmonicmean}
\end{center}
\end{figure}

\section{Conclusion}

    This experiment successfully demonstrated that combining fitness functions and RBMs it is possible autonomously detect faults and provide an accurate, ordered list of  potential root causes.
                It also expanded upon  prior research by demonstrating  better overall performance and the ability to predict a series of behaviours, but with the added caveats of requiring more time to produce  results from when the fault was initially detected, and higher variability.

Although the results from this experiment were positive, several questions remain unanswered--including whether or not this approach would reduce operating costs in a large-scale production environment, if using a larger data set would reduce variability, and if these approaches could be used to understand the source of a fault through feature locality.

Leveraging contrastive divergence learning in RBMs, makes it possible to  predict a sequence of feature behaviours.
This has several implications, but most notably 
using a multi-step prediction algorithm means there is potential to switch from a reactive to a pro-active detection of faults, and to understand if 
the root cause of a fault has multiple sources--(\textit{i.e.} feature locality).
By instantiating a self-adaptive primitive with the ability to predict a sequence of values, leveraging existing feature locality detection techniques becomes more accessible~\cite{garvin2013failureavoidanceinconfigurablesystemsthroughfeaturelocality,garvin2011usingfeaturelocality}.

Using a partially simulated data set has left some questions as to whether or not the results of this experiment could be improved further in terms of variability. Although this approach afforded a direct comparison to the prior research it would be interesting to see if doubling the training period would improve the results.
Similarly, it may be possible to train the RBM in a shorter amount of time by using a series of similarly configured virtual machines and then sharing the data between them over a network. One of the end goals in our research is to build a network aware, self-healing framework that is agnostic to its computing environment.

There are still other methodologies that   should be compared to better understand their advantages and disadvantages in anomaly detection.
More advanced types of neural network--such as long short term memory networks and  bio-directional recurrent neural networks (LSTMNs, BiRNNs, respectively). and other so-called `deep-belief' networks, represent interesting avenues for exploration.

It's important to note that WMI was not designed to provide the kind of functionality leveraged in this experiment. It does not have a primary key, nor a built-in mechanism for uniquely identifying rows of data--despite the fact that interfacing with WMI uses `WQL'. This poses challenges
when trying to determine if new items have been added, or existing items have been removed--such as a physical device or software application. It is for this reason that the dictionary with the unique identifier value was used. A replacement to WMI would provide substantial improvements to the speed  at which the data is gathered and compared, and promote more routine analysis in similar scenarios.

Lastly, some of the research avenues recommended in our previous work remain unexplored--including a live study of the self-healing systems frameworks in a large-scale computing environment, self-provisioning fitness tests, and understanding the differences in risk between supervised and unsupervised management techniques.

 \section*{Acknowledgments}

Funding for this research was provided by the Scottish Informatics and Computer
Science Alliance (SICSA). 

\bibliography{citations}{}
\bibliographystyle{ieeetran}

\end{document}